\documentclass{article}
\usepackage{ijcai20}

\usepackage{times}
\usepackage{soul}
\usepackage{url}
\usepackage[utf8]{inputenc}
\usepackage[small]{caption}
\usepackage{graphicx}
\usepackage{amsmath}
\usepackage{amsthm}
\usepackage{booktabs}
\usepackage{algorithm}
\usepackage{algorithmic}

\usepackage[hidelinks]{hyperref}
\newcommand{\fontsmall}{\fontsize{8pt}{8.5pt}\selectfont}

\newcommand{\fontmed}{\fontsize{8.5pt}{8.5pt}\selectfont}

\title{Natural Language QA Approaches using Reasoning with External Knowledge}

\author{
Chitta Baral$^1$\footnote{Contact Author}\and
Pratyay Banerjee$^1$\and
Kuntal Pal$^{1}$\And
Arindam Mitra$^2$\\
\affiliations
$^1$Arizona State University\\
$^2$Microsoft Inc.\\
\emails
\{chitta,pbanerj6,kkpal\}@asu.edu,
arindam.mitra@microsoft.com
}

\begin{document}

\maketitle

\begin{abstract}
Question answering (QA) in natural language (NL) has been an important aspect of AI from its early days. Winograd's ``councilmen'' example in his 1972 paper and McCarthy's Mr. Hug example of 1976 highlight the role of external knowledge in NL understanding. 
While Machine Learning  has been the go-to approach in NL processing as well as NL question answering (NLQA) for the last 30 years, recently there has been an increasingly emphasized thread on NLQA where external knowledge plays an important role. The challenges inspired by Winograd's councilmen example,
and recent developments such as the Rebooting AI book, various NLQA datasets, research on knowledge acquisition in the NLQA context, and their use in various NLQA models have brought the issue of NLQA using ``reasoning'' with external knowledge to the forefront. In this paper we present a survey of the recent work on them. We believe our survey will help establish a bridge between multiple fields of AI, especially between (a) the traditional fields of knowledge representation and reasoning and (b) the field of NL understanding and NLQA.  
\end{abstract}

\section{Introduction and Motivation}
Understanding text often requires knowledge beyond what is explicitly stated in the text. While this was mentioned in early AI works \cite{mccarthy1990example}, with recent success in  traditional NLP tasks, many challenge datasets have been recently proposed that focus on hard NLU tasks. The Winograd challenge was proposed in 2012, bAbI
\cite{weston2015towards}
, and an array of datasets were proposed in 2018-19. This includes\footnote{The website https://quantumstat.com/dataset/dataset.html 
has a large list of NLP and QA datasets. The recent survey \cite{storks2019recent} also mentions many of the datasets.}
QASC 
\cite{khot2019qasc}, 
CoQA
\cite{reddy2019coqa}
,
DROP 
\cite{Dua2019DROPAR}
, 
BoolQ
\cite{clark2019boolq}
, 
CODAH
\cite{chen2019codah}
, 
ComQA
\cite{abujabal2018comqa}
, 
CosmosQA
\cite{huang2019cosmos}
, 
NaturalQuestions 
\cite{kwiatkowski2019natural}
,
PhysicalIQA 
\cite{bisk2019piqa}
, 
QuaRTz
\cite{tafjord2019quartz}
,
QuoRef
\cite{dasigi2019quoref}
,
SocialIQA
\cite{sap2019socialiqa}
,
WinoGrande 
\cite{sakaguchi2019winogrande}
in 2019 and ARC
\cite{Clark2019FromT}
, 
CommonsenseQA 
\cite{talmor2018commonsenseqa}
,
ComplexWebQuestions
\cite{talmor2018web}
,
HotpotQA
\cite{yang2018hotpotqa}
,
OBQA
\cite{OpenBookQA2018}
,
Propara
\cite{mishra2018tracking}
,
QuaRel
\cite{tafjord2018quarel}
,
QuAC
\cite{choi2018quac}
,
MultiRC
\cite{MultiRC2018}
,
Record
\cite{zhang2018record}
,
Qangaroo
\cite{welbl2018constructing}
,
ShARC
\cite{sharc}
,
SWAG
\cite{zellers2018swag}
,
SQuADv2
\cite{rajpurkar2018know} 
in 2018.
%

%
To understand the technical challenges and nuances in building natural language understanding (NLU) and QA systems we consider a few examples from the various datasets. In his 1972 paper Winograd \cite{winograd1972understanding} presented the following ``councilmen'' example.   

\smallskip
\noindent
\fbox{
\begin{minipage}{0.95\linewidth}
\fontsmall
    \textbf{WSC item:} The city councilmen refused the demonstrators a permit because they [feared/advocated] violence.\\ \textbf{Question:} Who [feared/advocated] violence? 
\end{minipage}
}

\smallskip
In this example to understand what ``they'' refers to one needs knowledge about the action of ``refusing a permit to demonstrate''; when such an action can happen and with respect to whom?

The bAbI domain is a collection of 20 tasks and one of the harder task sub-domain in bAbI is the Path finding sub-domain. An example from that sub-domain is as follows:

\smallskip
\noindent
\fbox{
\begin{minipage}{0.95\linewidth}
\fontsmall
    \textbf{Context:} The office is east of the hallway. The kitchen is north of the office. The garden is west of the bedroom. The office is west of the garden. The bathroom is north of the garden.  \\
    \textbf{Question:} How do you go from the kitchen to the garden?\\
    \textbf{Answer:} South, East.
\end{minipage}
}

\smallskip
To answer the above question one needs knowledge about directions and their opposites, knowledge about the effect of actions of going in specific directions, and knowledge about composing actions (i.e., planning) to achieve a goal. 

Processes are actions with duration. Reasoning about processes, where they occur and what they change is somewhat more complex. The ProPara dataset \cite{mishra2018tracking} consists of natural text about processes. Following is an example.

\smallskip
\noindent
\fbox{
	\begin{minipage}{0.95\linewidth}
\fontsmall
\textbf{Passage:} Chloroplasts in the leaf of the plant trap light
from the sun. The roots absorb water and minerals from the soil. This combination of water and minerals flows from the stem into the leaf. Carbon dioxide (CO$_2$) enters the leaf. Light, water and minerals, and CO$_2$ all combine into a mixture. This mixture forms sugar
(glucose) which is what the plant eats.\\
\textbf{Question}: Where is sugar produced?
\textbf{Answer}: in the leaf
\end{minipage}
}

\smallskip
The knowledge needed in many question answering domain can be present in unstructured textual form. But often specific words and phrases in those texts have ``deeper meaning'' that may not be easy to learn from examples.  
Such a situation arises in the LifeCycleQA domain \cite{mitra2019declarative} where the description of a life cycle is given and questions are asked about them. One LifeCycleQA domain has the description of the life cycle of frog with five different stages: egg, tadpole, tadpole with legs, froglet and adult. It has description about each of these stages such as:

\smallskip
\noindent
\fbox{
\begin{minipage}{0.95\linewidth}
\fontsmall
    \textbf{Lifecycle Description:} Froglet - In this stage, the almost mature frog breathes with lungs and still has some of its tail. Adult - The adult frog breathes with lungs and has no tail (it has been absorbed by the body).  \\
    \textbf{Question:}  What best indicates that a frog has reached the adult stage? 
    \textbf{Options:} A) When it has lungs B) \textit{When the tail is absorbed by the body}
\end{minipage}
}

\smallskip
To answer this question one needs a precise understanding (or definition) of the word ``indicates''. In this case, (A) indicates that a frog has reached the adult stage (B) does not. That is because (B) is true for both adult and froglet stages while (A) is only true for the adult stage. The precise definition of ``indicates'' is that a property P indicates a stage $S_i$ with respect to a set of stages $\{S_1, \ldots, S_n\}$ iff P is true in $S_i$, and P is false in all stages in $\{S_1, \ldots, S_{i-1}, S_{i+1}, \ldots S_n\}$.

There are some datasets where it is explicitly stated that external knowledge is needed to answer questions in those datasets. Example of such datasets are the OpenBookQA dataset \cite{OpenBookQA2018} and the Machine Commonsense datasets \cite{sap2019socialiqa,bisk2019piqa,bhagavatula2019abductive}. Each OpenBookQA item has a question and four answer choices. An open book of science facts are also provided. In addition QA systems are expected to use additional common knowledge about the domain. Following is an example item from OpenBookQA.

\smallskip
\noindent
\fbox{
\begin{minipage}{0.95\linewidth}
\fontsmall
    \textbf{OpenBook Question:} In which of these would the most heat travel?\\
    \textbf{Options:} A) a new pair of jeans.
B) \textit{a steel spoon in a cafeteria}.
C) a cotton candy at a store.
D) a calvin klein cotton hat.\\
\textbf{Science  Fact:}  Metal is a thermal conductor\\
\textbf{Common Knowledge:} Steel is made of metal.
Heat travels through a thermal conductor.
\end{minipage}
}

\smallskip
As part of the DARPA MCS (machine commonsense) program Allen AI has developed 5 different QA datasets where reasoning with common sense knowledge (which are not given) is required to answer questions correctly. One of those datasets is the Physical QA dataset and following is an example from that dataset.

\smallskip
\noindent
\fbox{
\begin{minipage}{0.95\linewidth}
\fontsmall
\textbf{Physical QA:} You need to break a window. Which object would you rather use?\\
\textbf{Answer Options:}
A) \textit{metal stool}
B) bottle of water.
\end{minipage}
}

\smallskip
To answer the above question one needs knowledge about things that can break a window.

	One of the most challenging natural language QA task is solving grid puzzles \cite{mitra2015learning}. They contain a textual description of the puzzle and questions about the solution of the puzzle.  
Such puzzles are currently found in exams such as LSAT and GMAT, and have also been used to evaluate constraint satisfaction algorithms and systems.  Building a system to solve them is a challenge as it requires precise understanding of several clues of the puzzle, leaving very little room for error. In addition there is often need for external knowledge (including commonsense knowledge) that is not explicitly specified in the puzzle description.  

\smallskip
\noindent
\fbox{
\begin{minipage}{0.95\linewidth}
\fontsmall
    \textbf{Puzzle Context:} Waterford Spa had a full appointment calendar booked today. Help Janice figure out the schedule by matching each masseuse to her client, and determine the total price for each.\\
    \textbf{Puzzle Conditions:} Hannah paid more than Teri’s client. Freda paid 20 dollars more than Lynda’s client. Hannah paid 10 dollars less than Nancy’s client.  Nancy’s client, Hannah and Ginger were all different clients. Hannah was either the person who paid \$180 or Lynda’s client.
\end{minipage}
}


\smallskip 
The above examples are a small sample of examples from recent NLQA datasets that illustrate the need for ``reasoning'' with external knowledge in NLQA. In the rest of the paper we present several aspects of building NLQA systems for such datasets and give a brief survey of the existing methods. We start with a brief description of the knowledge repositories used by such NLQA systems and how they were created. We then present models and architectures of systems that do NLQA with external knowledge.

\section{Knowledge Repositories and their creation}
The knowledge repositories contain two kinds of knowledge: Unstructured and Structured Knowledge. Unstructured knowledge is knowledge in the form of free text or natural language. Structured knowledge, on the other hand, has a well-defined form, such as facts in the form of tuples or Resource Description Framework (RDF); or well-defined rules as used in Answer Set Program Knowledge Bases.
Recently pretrained language models are also being used as ``knowledge bases''. We discuss them in a later section.


{\bf Repositories of Unstructured Knowledge}: 
Any natural language text or book or even the web can be thought of as a source of unstructured knowledge. Two large repositories that are commonly used are the Wikipedia Corpus and the Toronto BookCorpus. Wikipedia corpus contains 4.4M articles about varied fields and is crowd-curated. Toronto BookCorpus consists of 11K books on various topics. Both of these are used in several latest neural language models to learn word representations and NLU.

A few other notable sources of unstructured commonsense knowledge are the Aristo Reasoning Challenge (ARC) Corpus \cite{Clark2019FromT}, the WikiHow Text Summarization dataset \cite{koupaee2018wikihow}, and the short stories from the RoCStories and Story Cloze Task datasets \cite{mostafazadeh2016corpus}. ARC Corpus contains 14M science sentences which are useful for the corresponding ARC Science QA task. WikiHow dataset contains 230K articles and summaries extracted from the online WikiHow website. This contains articles written by several different human authors, which is different from the news articles present in the Penn Treebank. RocStories and the Story Cloze dataset contains short stories that have a ``rich causal and temporal commonsense relations between daily events''. 

{\bf Repositories of Structured Knowledge}: 
There are several structured knowledge repositories, such as Yago
\cite{rebele2016yago}
, NELL
\cite{carlson2010toward}
, DBPedia
\cite{dbpedia}
and ConceptNet
\cite{liu2004conceptnet}
. Out of these, Yago is human-verified with the confidence of 95\% accuracy in its content. NELL on the other hand, continuously learns and updates new facts into its knowledge base. DBPedia is the structured variant of Wikipedia, with facts and relations present in RDF format. ConceptNet contains commonsense data and is an amalgamation of multiple sources, which are DBPedia, Wiktionary, OpenCyc to name a few. OpenCyc is the open-source version of Cyc, which is a long-living AI project to build a comprehensive ontology, knowledge base and commonsense reasoning engine. Wiktionary
\cite{wiki}
is a multilingual dictionary describing words using definitions and descriptions with examples. WordNet
\cite{fellbaum2012wordnet}
contains much more lexical information compared to Wiktionary, including words being grouped into sets of cognitive synonyms expressing a distinct concept. These sets are interlinked by conceptual-semantic and lexical relations.

ATOMIC
\cite{sap2019atomic}
, VerbPhysics
\cite{forbes2017verb}
and WebChild 
\cite{tandon2014webchild}
are recent collections of commonsense knowledge. WebChild is automatically created by extracting and disambiguating Web contents into triples of nouns, adjectives and relations like \emph{hasShape}, \emph{evokesEmotion}. VerbPhysics contains knowledge about physical phenomena. ATOMIC contains commonsense knowledge, that is curated from human annotators and focuses on inferential knowledge about a given event, such as, intentions of actors, attributes of actors, effects on actors, pre-existing conditions for an event and reactions of actors. It also contains relations, such as, \emph{If-Event-Then-Event}, \emph{If-Event-Then-MentalState} and \emph{If-Event-Then-Persona}. 


\section{Reasoning with external knowledge: Models and Architectures}
The first step in developing a QA system that incorporates reasoning with external knowledge is to have one or more ways to get the external knowledge. 

\subsection{Extracting the External Knowledge}

{\bf Neural Language Models:} Large neural models, such as BERT \cite{devlin2018bert}, that are trained on Masked Language Modeling (MLM) 
seem to encapsulate a large volume of external knowledge. Their use led to strong performance in multiple datasets and the tasks in GLUE leader board \cite{wang-etal-2018-glue}.
Recently, methods have been proposed that can extract structured knowledge from such models and are being used for relation-extraction and entity-extraction by modifying MLM as a ``fill-in-the-blank'' task \cite{petroni2019language,bouraoui2019inducing,DBLP:journals/corr/abs-1906-05317}. These neural models can also be used to generate sentence vector representations, which are used to find sentence similarity and knowledge ranking. 

{\bf Word Vectors:}
Even earlier word vectors such as Word2Vec \cite{mikolov2013distributed} and Glove \cite{pennington2014glove}
were used for extracting knowledge.
These word vectors imbibe knowledge about word similarities. These word similarities could be used for creating rules that can be used with other handcrafted rules in a symbolic logic reasoning framework. For example, \cite{beltagy-etal-2014-probabilistic} uses word similarities from word vectors and weighted rules in Probabilistic Soft Logic (PSL) to solve the task of Semantic Textual Similarity.

{\bf Information/Knowledge Retrieval:}  However, the task of open-domain NLQA, such as in OpenBookQA mentioned earlier, need external unstructured knowledge in the form of free text.
The text is first processed using techniques aimed at removing noise and making retrieval more accurate, such as removal of punctuation and stop-words, lower-casing of words and lemmatization. The processed text is then stored in reverse-indexed in-memory storage such as Lucene \cite{10.5555/1893016} and retrieval servers like ElasticSearch \cite{gormley2015elasticsearch}. The knowledge retrieval step consists of search keyword identification, initial retrieval from the search engine and then depending on the task, a knowledge re-ranking step. Some complex tasks which require multi-hop reasoning, perform additional such knowledge retrieval steps. In these tasks, the keyword identification step is modified to account for the knowledge sentences retrieved in the previous steps. Multiple such queries make the task computationally expensive.

Few of the structured knowledge sources such as ATOMIC and ConceptNet also contain natural language text examples. These are used as unstructured knowledge and retrieved using the above techniques.

{\bf Semantic Knowledge Ranking/Retrieval:} The knowledge sentences retrieved through IR are re-ranked further using Semantic Knowledge Ranking/Retrieval (SKR) models, such as in \cite{pirtoaca2019answering,banerjee-etal-2019-careful,mitra2019exploring,banerjee-2019-asu}.
In SKR, instead of traditional retrieval search engines, neural networks are used to rank knowledge sentences. These neural networks are trained on the task of semantic textual similarity (STS), knowledge relevance classification or natural language inference (NLI).


{\bf Existing structured knowledge sources:}
ConceptNet, WordNet and DBPedia are examples of structured knowledge bases, where knowledge is stored in a well-defined schema. This schema allows for more precise knowledge retrieval systems. On the other hand, the knowledge query too, needs to be well defined and precise. This is the challenge for knowledge retrieval from structured knowledge bases.
In these systems, the question is parsed to a well-defined structured query that follows the strict semantics of the target knowledge retrieval system. For example, DBPedia has knowledge stored in RDF triples. A question parser translates a natural language question to a SPARQL query which is executed by the DBPedia Server.


{\bf Hand coded knowledge:} However, in domains such as the LifeCycleQA where precise definitions of certain terms are needed, such knowledge is hand coded. Similarly, in the puzzle solving domain, ASP rules encoding the fundamental assumption about such puzzles  are hand-crafted. One such assumption is that, associations between the elements are unique. The ASP rules for this is:

\noindent
{ \centering \fontsmall 1 \{tuple(G,Cat,Elem): element(Cat,Elem) \} 1 :-  cindex(Cat), eindex(G).\\
\hspace{5pt} :- tuple(G1,Cat,Elem), tuple(G2,Cat,Elem),G1!=G2.}
\smallskip

{\bf Knowledge learned using Inductive Logic Programming (ILP)}: Sometimes logical ASP rules can be learned from the dataset. Such a method is used in \cite{mitra2019declarative,mitra2019knowledge} to learn rules that are used to solve bAbI, ProPara and Math word problem datasets. There, ILP (inductive logic programming) is used  to  learn  effect  of  actions, and static  causal  connection  between properties of the world. Following is an example of a rule that is learned.

\noindent 
{
\fontsmall
\hspace{5pt} initiatedAt(carry(A, O), T ) :- happensAt(take(A, O), T ).
}

\noindent
The rule says that  A is carrying O gets initiated at time point T if the event take(A,O) occurs at T.


\subsection{Models and architecture for NLQA with external knowledge}

\begin{figure}
\small
  \includegraphics[width=\linewidth,height=0.5\linewidth]{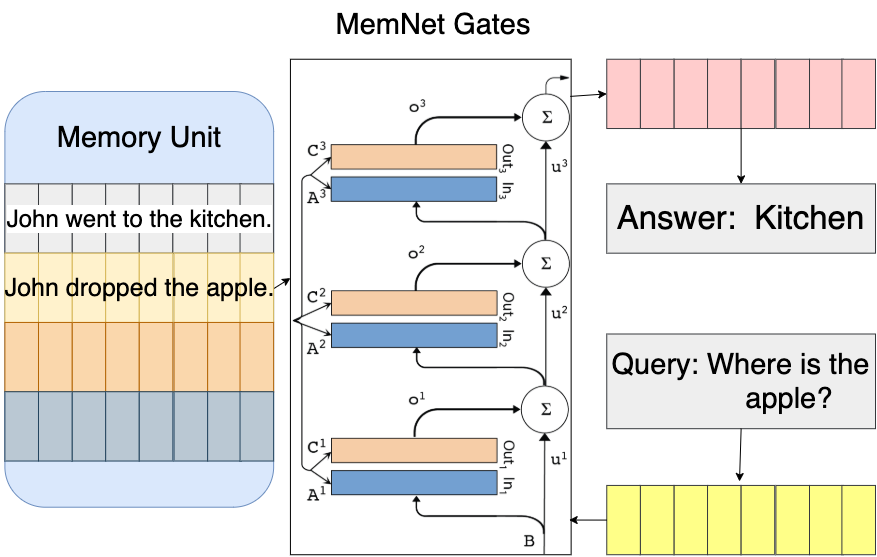}
  \caption{Memory Network with Free text Knowledge. This is an example from the bAbI dataset. The memory unit is first updated from reading the knowledge passage and finally a query is given.}
  \label{fig:memnet}
\end{figure}

NLQA systems with external knowledge can be grouped based on how  knowledge is expressed (structured, free text, implicit in pre-trained neural networks, or a combination) and the type of reasoning module (symbolic, neural or mixed).

\smallskip 
\noindent
\fbox{
\begin{minipage}{0.95\linewidth}
\fontmed
\textbf{Answer Set Programs:} 
An answer set program (ASP) is a collection of rules of the form,
$L_0 \leftarrow L_1, \ldots , L_m, \ not \  L_{m+1}, ..., \ not \  L_n$,
where each of the $L_i$ is a literal in the sense of a classical
logic. Intuitively, the above rule means that if $L_1, \ldots , L_m$ are
true and if $L_{m+1}, \ldots, L_n$ can be safely assumed to be false
then $L_0$ must be true.  The semantics of ASP is based on the stable model (answer set) semantics of logic programming. ASP is one of the declarative knowledge representation and reasoning language with a large 
building block results, efficient interpreters and large variety of applications. 
\end{minipage}
}

\smallskip 

{\bf Structured Knowledge and Symbolic Reasoner}: 
While very early NLQA systems were symbolic systems, and most recent NLQA systems are neural, there are a few symbolic systems that do well on the recent datasets. 
As mentioned earlier the system \cite{beltagy-etal-2014-probabilistic} uses probabilistic soft logic to address semantic textual similarity. The system in \cite{mitra2016addressing} learns logical rules from the bAbI dataset using ILP and then does reasoning using ASP and performs well on that dataset. 

\smallskip 
\noindent
\fbox{
\begin{minipage}{0.95\linewidth}
\fontmed
\textbf{Inductive Logic Programming:} 
Inductive Logic Programming (ILP)  is
a subfield of Machine learning that is focused on learning
logic programs. Given a set of positive examples $E^+$, negative 
examples $E^-$ and some background knowledge
B, an ILP algorithm finds an Hypothesis H (answer set
program) such that $B \cup H \models E^+$ 
and $B \cup H \not \models E^-$. The possible hypothesis space is often restricted with a language bias that is specified by a series of mode
declarations. Recently, ILP methods have been developed with respect to Answer Set Programming.
\end{minipage}
}

\smallskip

The system \cite{mitra2019knowledge} addresses the Propara dataset in a similar manner and performs well. The system to solve puzzles is also symbolic and so far there has not been any neural system addressing this dataset. However none of these systems are end-to-end symbolic systems as they use neural models for tasks (in the overall pipeline) such as semantic parsing of text.

\smallskip
\noindent
\fbox{
\begin{minipage}{0.95\linewidth}
\fontmed
\textbf{Transformers:} 
Generative Pre-trained Transformer(GPT) \cite{radford2018improving} showed that,  pretraining on diverse unstructured  knowledge texts using transformers \cite{vaswani2017attention} and then selectively fine-tune on specific tasks, helps achieving  state-of-the-art systems. Their unidirectional way learning representations from diverse texts was improved by bidirectional approach in BERT. It uses the transformer encoder as part of their architecture and takes a fixed length of  input and every token is able to attend to both its previous and next tokens through the self attention layers of the transformers.  
Robustly optimized BERT approach, RoBERTa \cite{liu2019roberta} is another variant  which uses a different pre-training approach. RoBERTa is trained on ten times more data than BERT and it outperforms BERT in almost all the NLP tasks and benchmarks. 
\end{minipage}
}
\smallskip

{\bf Neural Implicit Knowledge with Neural Reasoners:} Neural network models based on transformers are able to use the knowledge embedded in their learned parameters in various downstream NLP tasks. These models learn the parameters as a part of \textit{pre-training} phase on a huge collection of diverse free texts. They fine-tune the parameters  for selective NLP tasks and in the process refine the pre-trained knowledge even further for a domain and a task. This knowledge in the parameters of the models helps then in 
multiple NLP tasks. BERT learns the knowledge from mere 16GB, while Roberta uses 160GB of free texts for  pre-training.
We can say that the knowledge learned during the pre-training phase is generic for any domain and for any NLP task which gets tuned for a particular domain and task during the fine-tuning phase.

{\bf Structured Knowledge and Neural Reasoners:} Structured knowledge  can be in the form of trees (abstract syntax tree, dependency tree, constituency tree), graphs, concepts or rules.
Tree-structured knowledge extracted from input text have been used in the Tree-based LSTM \cite{tai2015improved}. Here, aggregated knowledge from multiple child nodes selected through the gating mechanism in the model  propagates to the parent nodes and creates a representation of the whole tree and thus improving the downstream NLP tasks.

\smallskip
\noindent
\fbox{
\begin{minipage}{0.95\linewidth}
\fontmed
\textbf{LSTMs:} 
Long-Short Term Memory  \cite{hochreiter1997long} is a type of Recurrent Neural Networks that are used to deal with variable-length sequence inputs. 
Reasoning and QA with LSTMs are usually modeled as an encoder-classifier architecture. The encoder network is used to learn a vector representation of the question and external knowledge. The classifier network either classifies word-spans to extract answer or answer options to in a MCQA setting. These models are able to encode variable-length sentences and hence can take as input a considerable amount of external knowledge. The limitations of LSTMs are that they are computationally expensive and have limited memory, i.e,  they can only remember last  $T$ time-steps.
\end{minipage}
}
\smallskip

\begin{figure}
  \includegraphics[width=\linewidth,height=0.5\linewidth]{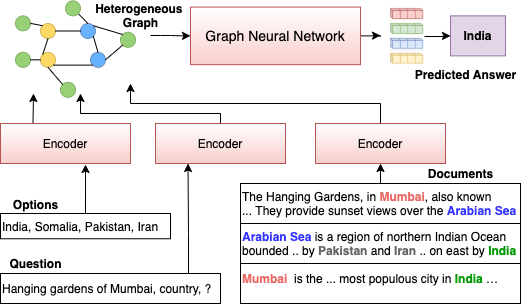}
  \caption{Heterogeneous Graph Neural Network with Structured Knowledge. The example is taken from WikiHop dataset. Based on the documents, question and answer options a heterogeneous graph is created(document nodes in green, option nodes in yellow and entity nodes in blue)}
  \label{fig:gnn}
\end{figure}

While knowledge in the form of undirected graphs are mainly used by the graph-based reasoning systems like Graph Neural Networks (GNN) \cite{scarselli2008graph}, the directed graphs are better handled by  convolutional neural networks  with dense  graph propagation  \cite{kampffmeyer2019rethinking}. If the  knowledge is heterogeneous in nature, having multiple types of objects(graph nodes) and the edges carry different semantic information,  heterogeneous graph attention networks 
\cite{wang2019heterogeneous} 
is a common choice. Figure \ref{fig:gnn} shows an example. Knowledge carrying information in hierarchy (entity-level, sentence-level, paragraph-level and document-level) makes use of Hierarchical Graph neural network \cite{fang2019hierarchical} for better performance.


\noindent
\fbox{
\begin{minipage}{0.95\linewidth}
\fontmed
\textbf{Graph Neural Networks:} 
GNNs \cite{scarselli2008graph} have gained immense importance recently because of it higher expressive power and generalizability across various domains.  GNN learns the representation of each nodes of the input graphs by aggregating information from its neighbors through message passing \cite{gilmer2017neural}. It learns both semantically and structurally rich representation by iterating over its neighbors in successive hops. 
 In spite of their heavy usage across multiple domains, their shallow structure(in terms of the layers of GNN), inability to handle dynamic graphs and scalability are still open problems \cite{zhou2018graph}.
 
\end{minipage}
}

\smallskip
Knowledge can be in the form of structured concepts. One such approach was taken in Knowledge-Powered convolutional neural networks
\cite{wang2017combining} 
where they use concepts from Probase
to classify short text.
In another approach, K-BERT 
\cite{liu2019k} 
inject knowledge graph triples along with other text and together learn representation which helps in NLQA and named entity recognition tasks.

Structured knowledge can be in the form of rules which are compiled into the neural network. One such approach \cite{hu-etal-2016-harnessing} was taken to incorporate first order logic rules into the neural network parameters.

Structured knowledge can also be embedded into a neural network through synthetically created training samples.  This can be either template-based or specially hand-crafted to exploit a particular behavior of any model. The knowledge bases such as ConceptNet, WordNet, DBPedia are used to create these hand-crafted training examples. For example, in the NLI task, new knowledge is hand-crafted from these knowledge bases by normalizing the names and switching the roles of the actors. This augmented knowledge helped in differentiating between similar sentences through a symbolic attention mechanism in neural network
\cite{mitra2019understanding}.

{\bf Free text Knowledge and Neural Reasoners:} 
Memory networks have been used in prior work on tasks which need long term memory such as the tasks in bAbI datasets \cite{kumar2016ask,liu2017gated}. In these models, free text is used as external knowledge and stored in the memory units. The answering module uses this stored memory representation to reason and answer questions.

\smallskip
\noindent
\fbox{
\begin{minipage}{0.95\linewidth}
\fontmed
\textbf{Memory Networks:} 
Memory Networks augment neural networks with a read and writable \emph{memory} section. These networks are designed to read natural language text and learn to store these input to a specially designed \emph{memory} in which the free text input or a vector representation of the input is written under certain conditions defined by gates. At inference time, a read operation is performed on the memory. These networks are made robust to unseen words during inference with techniques of approximating word vector representations from neighbouring words. These networks outperform RNNs and LSTMs on tasks which require long term memory utilizing both the dedicated \emph{memory} unit and unseen word vectors. Their major limitation lies in their difficulty to be trainable using back-propagation and requiring full supervision during training.
\end{minipage}
}
\smallskip

Free text Knowledge in the form of topics, tags and entities have been used along with texts to improve the QA tasks using a knowledge enhanced hybrid neural network \cite{wu2018knowledge}. They used knowledge gates to match semantic information of the input text with the knowledge weeding out the unrelated texts.

\begin{figure}
  \includegraphics[width=\linewidth,height=0.5\linewidth]{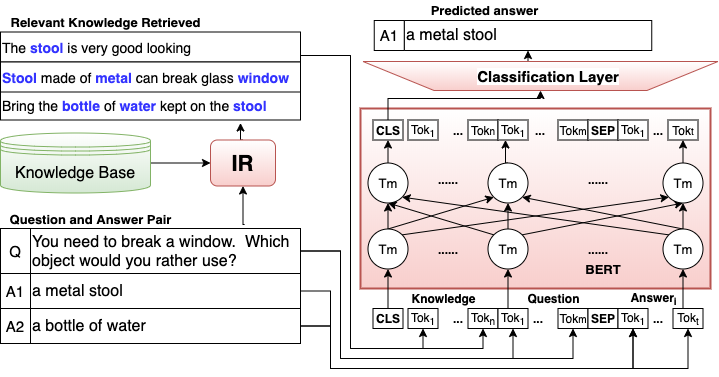}
	  \caption{An example from OBQA. Transformer based Neural Models with Free text Knowledge obtained by Information Retrieval on Knowledge Bases. }
  \label{fig:trans_free}
\end{figure}

In the scenario where the knowledge embedded in parameters of neural models are not enough for an NLP task, external knowledge is often infused in the free text forms along with the model inputs. This has often led to improved reasoning systems. 
Commonsense reasoning is one such area where external knowledge in the free text form helps in achieving better performance. Multiple challenges has been mentioned earlier which require reasoning abilities.

A common approach  for using the free text along with the reasoning systems can be seen from Figure \ref{fig:trans_free}. The idea is to pre-select  a set of appropriate knowledge texts (using IR) from an unstructured knowledge repository or document and then let the previously learned reasoning models use it \cite{banerjee-etal-2019-careful,clark2018think,mitra2019understanding}.  




{\bf Free text Knowledge and Mixed Reasoners:}
There have been few approaches in bringing the neural and symbolic approaches together for commonsense reasoning with knowledge represented in text. 
One such neuro-symbolic approach NeRD \cite{chen2019neural} was used to achieve state-of-the-art performance in two mathematical datasets DROP and MathQA. NeRD consists of a \textit{reader} which generates representation from questions with passage and a \textit{programmer} which generates the reasoning steps that when executed produces the desired results.

In the LifeCycleQA dataset terms with deeper meaning, such as ``indicates'',  are defined explicitly using 
Answer Set Programming (ASP). The overall reasoning system in \cite{mitra2019declarative} is an ASP program that makes use of such ASP defined terms and performs external calls to a neural NLI system. In \cite{prakash2019combining} the Winograd dataset is addressed by using ``extracted sentences from web'' and 
BERT
and combined reasoning is done using PSL. 

{\bf Combination of Knowledge and Mixed Reasoners:}
The ARISTO solver by AllenAI \cite{Clark2019FromT} consists of eight reasoners that utilize a combination of external knowledge sources. The external knowledge sources include Open-IE \cite{etzioni2008open}, the ARC corpus, TupleKB, Tablestore and TupleInfKB \cite{clark2018think}. The solvers range from retrieval-based and statistical models, symbolic solvers using ILP, to large neural language models such as BERT. They combine the models using two-step ensembling using logistic regression  over base reasoner scores.

\section{Discussion: How much Reasoning the models are doing?}
While it is easy to see the various kinds of reasoning done by symbolic models, it is a challenge to figure out how much reasoning is being done in neural models. The success with respect to datasets  give some indirect evidence, but more thorough studies are needed. 

\noindent
\paragraph{Commonsense Reasoning:}
The commonsense datasets, (CommonsenseQA, CosmosQA, SocialIQA, etc.), require commonsense knowledge to be solved. So the systems solving them should perform deductive reasoning with external commonsense knowledge. BERT-based neural models are able to perform reasonably well on such tasks, with accuracy ranging from 75\%-82\% which falls quite short of human performance of 90-94\% \cite{mitra2019exploring}. They leverage both external commonsense knowledge and knowledge learned through pre-training tasks. Inability of the pure neural models to represent such a huge variety of commonsense knowledge as rules led to pure symbolic or neuro-symbolic approaches on such datasets. 

\noindent
\paragraph{Multi-Hop Reasoning:}
NLQA datasets such as, Natural Questions, HotpotQA, OpenBookQA, etc, need systems to combine information spread across multiple sentences, i.e, Multi-Hop Reasoning.
Current state-of-the-art neural models are able to achieve performance around 65-75\% on these tasks \cite{banerjee-etal-2019-careful}. Transformer models have shown a considerable ability to perform multi-hop reasoning, particularly for the case where the entire knowledge passage can be given as an input, without truncation. Systems have been designed to select and filter sentences to reduce confusion in neural models
\noindent
\paragraph{Abductive Reasoning:}
AbductiveNLI, a dataset from AllenAI, defines a task to abduce and infer which of the possible scenarios best explains the consequent events. 
Neural models have shown a considerable performance in the simple two-choice MCQ task \cite{mitra2019exploring}, but have a low performance in the abduced hypothesis generation task. This shows the neural models are able to perform abduction to a limited extent.

\noindent
\paragraph{Quantitative and Qualitative Reasoning:}
Datasets such as DROP, AQuA-RAT, Quarel, Quartz require systems to perform both qualitative and quantitative reasoning. 
The quantitative reasoning datasets like DROP, AQuA-RAT, led to emergence of approaches which can learn symbolic mathematics using deep learning and neuro-symbolic models \cite{Lample2020Deep,chen2019neural}.
Neural as well as symbolic approaches have shown decent performance(70-80\%) on qualitative reasoning compared to human performance of (94-95\%) \cite{tafjord2018quarel,tafjord2019quartz}.
All the datasets have shown a considerable diversity in systems with all three types (neural, symbolic and neuro-symbolic) of models performing well. 


\noindent
\paragraph{Non-Monotonic Reasoning:}
In \cite{clark2020transformers} an attempt is made to understand the limits of reasoning of Transformers. They note that the transformers can perform simple logical reasoning and can understand rules in natural language which corresponding to classical  logic and can perform limited non-monotonic reasoning. Though there are few datasets which require such capabilities. 



\section{Conclusion and Future Directions}
In this paper we have surveyed \footnote{\href{https://bit.ly/3836xAW}{A version of this paper with a larger bibliography is linked here.} If accepted, we plan to buy extra pages to accommodate them.} recent research on NLQA when external knowledge -- beyond what is given in the test part -- is needed in correctly answering the questions. We gave several motivating examples, mentioned several datasets, discussed available knowledge repositories and methods used in selecting needed knowledge from larger repositories, and analyzed and grouped several models and architectures of NLQA systems based on how the knowledge is expressed in them and the type of reasoning module used in them.  Although there have been some related recent surveys, such as \cite{storks2019recent}, none of them focus on how exactly knowledge and reasoning is done in the NLQA systems dealing with datasets that require external knowledge. Our survey touched upon knowledge types that include structured knowledge, textual knowledge, knowledge embedded in a neural network, knowledge provided via specially constructed examples and combinations of them. We explored how symbolic, neural and mixed models process and reason with such knowledge. Based on our observations for various models following are some questions and future directions.

Following up on the LifeCycleQA dataset where some concepts, such as ``indicates'', were manually defined, several questions -- with partial answers -- may come to mind. (i) How big is the list of such concepts? If a list of them is made and they are defined then these definitions can be directly used or compiled into neural models. Can Cyc and the book \cite{gordon2017formal} be starting points in this direction?  (ii) Can these definitions be learned from data? How? Unless one only focuses on specific datasets, the challenge in learning these definitions would be that for each of them specialized examples would have to be created. (iii) When is it easier to just write the definitions in a logical language? When is it easier to learn from data sets?  Some concepts are easy to define logically but may require lot of examples to teach a system. There are concepts whose definitions took decades for researchers to formalize. An example is the solutions of frame problem. So it is easier to use that formalization rather than learning from scratch. On the other hand the notion of ``cause'' is still being fine tuned. Another direction of future work centers around the question of how well neural network can do reasoning; What kind of reasoning they can do well and what are the challenges ?

We hope this paper encourages further interactions between researchers in the traditional area of knowledge representation and reasoning who mostly focus on structured knowledge, and researchers in NLP/NLU/QA who are interested in NLQA where reasoning with knowledge is important. That would be a next step towards better understanding of natural language and development of cognitive skills. It will help us move from knowledge and comprehension to application, analysis and synthesis, as described in Bloom's taxonomy.


\newpage

\bibliographystyle{plain}
\bibliography{full-updated}

\end{document}